\documentclass[conference,a4paper]{IEEEtran}
\IEEEoverridecommandlockouts
\usepackage{textcomp}
\usepackage{subfigure}
\usepackage{diagbox}
\usepackage{graphicx}
\usepackage{amsmath,amssymb} % define this before the line numbering.
\usepackage{color}
\usepackage{diagbox}
\usepackage{float}
\usepackage{subfigure}
\usepackage{url}
\usepackage{amsthm}

\def\BibTeX{{\rm B\kern-.05em{\sc i\kern-.025em b}\kern-.08em
    T\kern-.1667em\lower.7ex\hbox{E}\kern-.125emX}}

\title{DSAN: Double Supervised Network with Attention Mechanism for Scene Text Recognition}

\begin{document}

\author{Yuting Gao$^{\star}$ \quad Zheng Huang$^{\star \dagger}$\thanks{$\dagger$ Corresponding author: huang-zheng@sjtu.edu.cn} \quad Yuchen Dai$^{\star}$ \quad Cheng Xu$^{\star}$ \quad Kai Chen$^{\star}$ \quad Jie Guo$^{\star}$ \\
	$^{\star}$ Shanghai Jiao Tong University, Shanghai, China \\
	$^{\dagger}$ Westone Cryptologic Research Center, Beijing, China}

\maketitle

\begin{abstract}
In this paper, we propose Double Supervised Network with Attention Mechanism (DSAN), a novel end-to-end trainable framework for scene text recognition. It incorporates one text attention module during feature extraction which enforces the model to focus on text regions and the whole framework is supervised by two branches. One supervision comes from context-level modelling branch and another comes from one extra supervision enhancement branch which aims at tackling inexplicit semantic information at character level. These two supervisions can benefit each other and yield better performance. The proposed approach can recognize text in arbitrary length and does not need any predefined lexicon. Our method achieves the current state-of-the-art results on three text recognition benchmarks: IIIT5K, ICDAR2013 and SVT reaching accuracy 88.6\%, 92.3\% and 84.1\% respectively which suggests the effectiveness of the proposed method.
\end{abstract}


\section{Introduction}
Scene text contains rich and high-level semantic information which is crucial for scene understanding. Recognizing text in the wild usually involves two steps, text detection and text recognition. In this paper, we focus on text recognition which directly convert cropped text region images into text strings. Since the development of deep learning, scene text recognition has become a hot topic in computer vision. Although OCR has been quite successful, it mainly aims at neat document. Due to the complexity of natural scenes, scene text recognition is still challenging. Fig.1 shows some quite difficult examples.

The current mainstream scene text recognition method is to treat it as sequence recognition which usually employs RNN to capture contextual information. We call this approach context-level modelling. Although some texts in natural images do have explicit semantic information, such as English words, many scene texts just are simple concatenation of characters without explicit contextual relationship, such as Fig.1(c). At this point, just use context-level modelling usually does not handle decently.

In this paper, we present a novel and effective framework called Double Supervised Network with Attention Mechanism (DSAN) for lexicon-free scene text recognition. Different from existing methods, DSAN treats semantic information in text from two perspectives. It contains two supervision branches to tackle explicit and inexplicit semantic information respectively. Specifically, beyond context-level modelling branch which aims at capturing explicit semantic information, one supervision enhancement branch is applied to handle inexplicit semantic information at character level. These two branches have mutually reinforcing effects and they are jointly trained together. In addition, one text attention module is integrated in the entire framework which can make the model to adaptively focus on the most important areas. We conduct experiments on three benchmarks and the results demonstrate the effectiveness of the proposed method.

\begin{figure}[H]
	\vspace{-9pt}
	\centering
	\subfigure[]{
		\includegraphics[width=1.05in,height=0.38in]{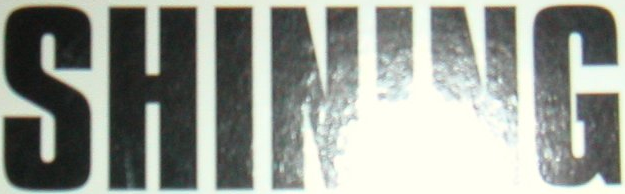}}
	\subfigure[]{
		\includegraphics[width=1.05in,height=0.38in]{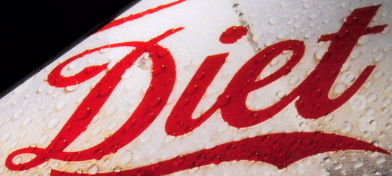}}
	\subfigure[]{
		\includegraphics[width=1.05in,height=0.38in]{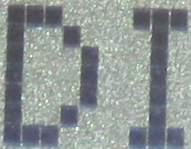}}
	\subfigure[]{
		\includegraphics[width=1.05in,height=0.38in]{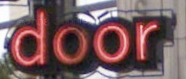}}
	\subfigure[]{
		\includegraphics[width=1.05in,height=0.38in]{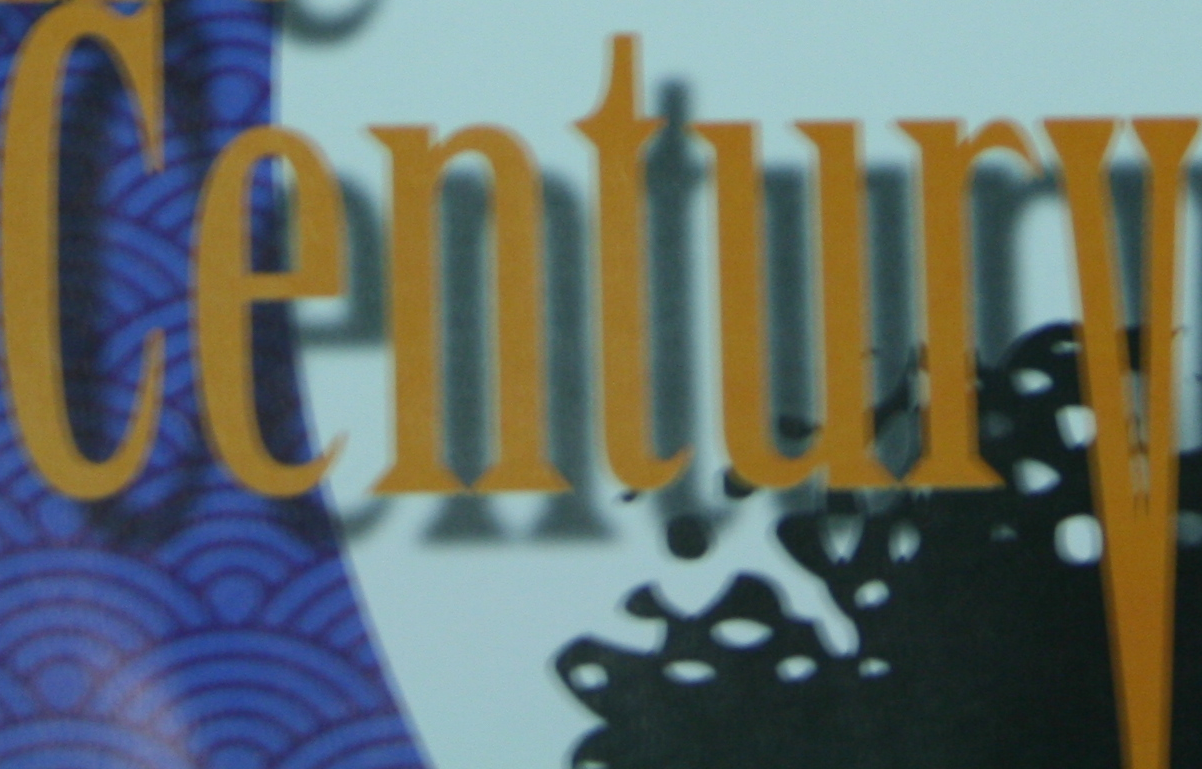}}
	\subfigure[]{
		\includegraphics[width=1.05in,height=0.38in]{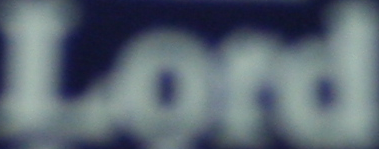}}
	\vspace{-5pt}
	\caption{Difficult examples. (a) uneven illumination (b)(c) variable fonts (d) complicated background (e) variable font sizes (f) blur}
	\vspace{-5pt}
\end{figure}

\begin{figure*}[htbp]
	\centering
	\includegraphics[height=2.55in,width=5.4in]{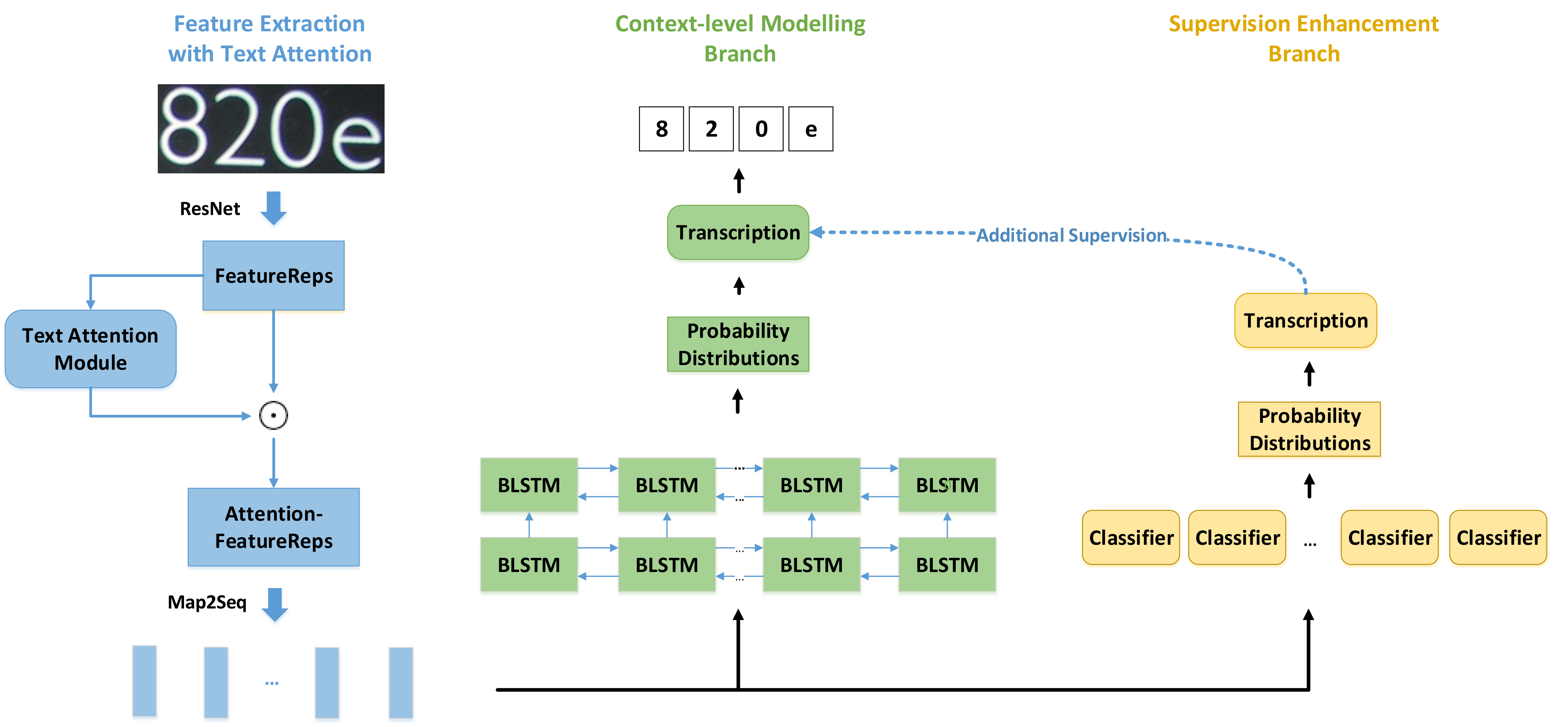}
	\vspace{-6pt}
	\caption{The proposed framework consists of three parts: text attention-based feature extraction, context-level modelling branch and one extra supervision enhancement branch. Extracted feature sequence will be input to two branches to do context-level modelling and character-level modelling respectively.} 
	\vspace{-12pt}
\end{figure*}

\vspace{-10pt}

\section{RELATED WORK}

Scene text recognition approaches can be divided into two categories, character-based and word-based. Traditional approaches are character-based which means they recognize text character by character\cite{Wang2013End}\cite{Bai2014Strokelets}. Due to the lack of contextual relationship between characters, it is difficult to recognize words totally correct at character level. Advanced studies treat text recognition as sequence recognition which can recognize word directly\cite{Shi2017An}\cite{Lee2016Recursive}\cite{Cheng2017Focusing}. 

Inspired by these existing methods, we present DSAN which can recognize text in arbitrary length. DSAN inherits merits from both traditional methods and advanced methods. It takes explicit and inexplicit semantic information into consideration simultaneously. 

\vspace{-4pt}

\section{METHODS}
The proposed framework is diagrammed in Fig.2. It is an end-to-end trainable network which consists of three parts, text attention-based feature extraction, context-level modelling branch and character-level modelling branch. Specifically, feature representations of each image are extracted through a ResNet-34 based backbone. Furthermore, one dedicated text attention module is integrated in the extraction procedure to make the model selectively focus on important parts. The final extracted features are then reshaped to feature sequence and input to two branches. One branch is Context-level Modelling Branch which uses two-layer bidirectional LSTM to capture contextual information. Another branch is Supervision Enhancement Branch which is composed of character classifiers and mainly aimed at tackling inexplicit semantic information at character level. These two branches are jointly trained together through a multi-task loss.

\subsection{Network Architecture}
\subsubsection{\textbf{Feature Extraction with Text Attention}}
Feature representations of each image are extracted through a ResNet-34 based backbone, denoted as FeatureReps. The stride of original ResNet-34 is 32, which is too large for cropped word images. Thus, we did some modifications to make it more appropriate. Specifically, we replace the 7x7 convolution with 3x3 same convolution and remove the 3x3 maxpooling. As a result, the height and width of FeatureReps both are $\frac{1}{8}$ of the input image and the number of channels is 512. 

After acquiring FeatureReps, text attention module is applied on it. As shown in Fig.3, one 3x1 with stride 1x1 same convolution is applied on FeatureReps and a sigmoid activation function follows. The output of text attention module is called AttentionMask which has the same resolution as FeatureReps but only one channel. Then weighted feature representations called AttentionFeatureReps are obtained through broadcast elementwise product operation (due to different channel numbers) on FeatureReps and AttentionMask.

\vspace{-9pt}
\begin{figure}[htbp]
	\centering
	\includegraphics[height=0.7in,width=3.4in]{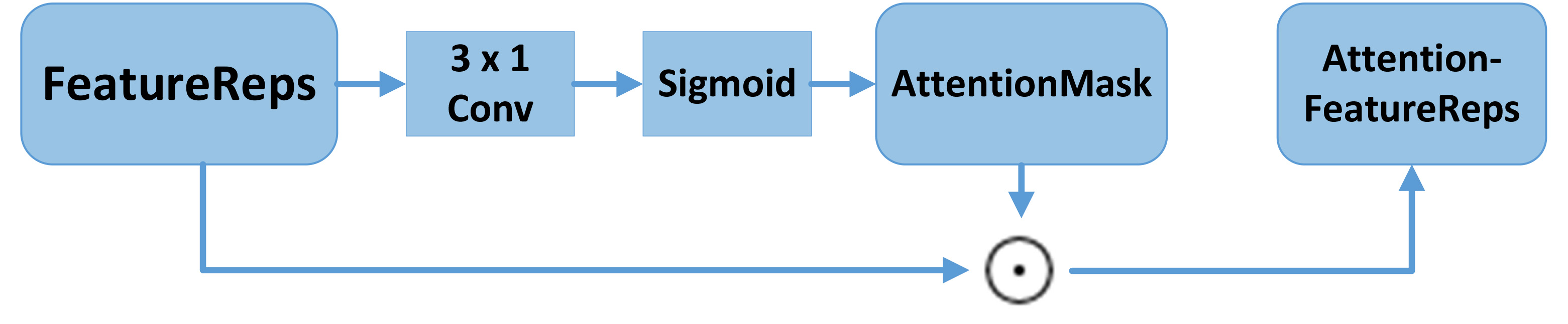}
	\vspace{-8pt}
	\caption{Text attention module. One 3x1 with stride 1x1 same convolution followed by a sigmoid activation function. The circle dot operation is broadcast elementwise product operation.}
	\vspace{-5pt}
\end{figure}

\vspace{-4pt}

In the text attention module, it actually does a feature filtering process, enhancing feature columns with semantic information while suppressing redundancies and clutters. Some heatmap examples of AttentionMask and its corresponding input images are shown in Fig.4. Darker regions obtain higher values which means these regions are highly activated. These heatmaps exactly demonstrate that the proposed text attention module do make the model focus on important parts. In Fig.4(a) the area between characters are low activated and in Fig.4(b) the edge portion of the image is selectively ignored. This module can mitigate the effects of inaccurate text detection results, such as the detected bounding box is not very close to the text area. Meanwhile, it is self-adaptive and does not need extra supervision information.

\vspace{-5pt}

\begin{figure}[htbp]
	\vspace{-4pt}
	\centering
	\subfigure[]{
		\begin{minipage}[t]{0.5\linewidth}
			\centering
			\includegraphics[height=0.42in,width=1.03\linewidth]{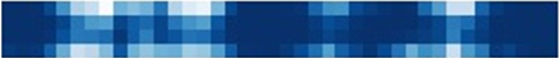}\\
			\vspace{6pt}
			\includegraphics[height=0.42in,width=1.03\linewidth]{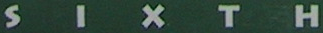}\\
			\vspace{2pt}
			%\centerline{(a)}
		\end{minipage}
	}
	\subfigure[]{
		\begin{minipage}[t]{0.4\linewidth}
			\centering
			\includegraphics[height=0.42in,width=0.95\linewidth]{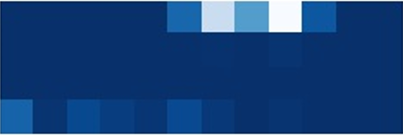}\\
			\vspace{6pt}
			\includegraphics[height=0.42in,width=0.95\linewidth]{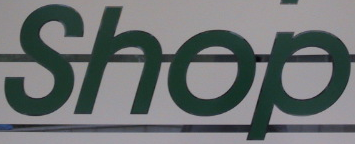}\\
			\vspace{2pt}
		\end{minipage}
	}
	\centering
	\vspace{-4pt}
	\caption{Heatmap examples of AttentionMask and corresponding input images.}
\end{figure}

\vspace{-5pt}

Finally, AttentionFeatureReps are converted to feature sequence. Suppose AttentionFeatureReps have the size of HxWxD, where H, W, D are height, width and depth respectively. Then Map-to-Sequence\cite{Shi2017An} is adopted to convert AttentionFeatureReps into sequence of W feature vectors, each vector has $HD$ dimensions and corresponds to one image region. The feature sequence will input to context-level modelling branch and supervision enhancement branch simultaneously.

\subsubsection{\textbf{Context-level Modelling Branch}}

Due to the high-level semantic information in text, context-level modelling is of great significance to text recognition. For text, contexts from both forward direction and backward direction are useful and important. Thus, the extracted feature sequence is fed into two-layer bidirectional LSTM to do context-level modelling which is effective to capture bidirectional long-term dependence in text. The output of BLSTM is $\textbf{h} = (h_1,h_2,...,h_W)$, where W is the length of input sequence. Then each vector in the output sequence will put into a fully connected layer and followed by a softmax classifier over the alphabet. 

This branch will generate probability distribution over the alphabet for each feature vector, then the probability distributions will input to transcription layer to generate output.  

\subsubsection{\textbf{Supervision Enhancement Branch}}

Some texts in natural scenes just are simple combinations of characters without explicit semantic information, just doing context-level modelling for these texts usually does not handle decently. Therefore, we propose supervision enhancement branch to tackle this problem which does character-level modelling for texts and aims at obtaining finer features for each character.

The extracted feature sequence is fed into a classification layer which is composed of character classifiers. Each feature vector 
is directly classified over the alphabet with a softmax classifier. Finally, the probability distributions are input to transcription layer. 

This branch does not consider contextual relationship between feature vectors, each feature vector is independent. It can make model extract finer feature for each character, thus feature sequence can be better constrained which is also advantageous for context-level modelling.

%Meanwhile, context-level modelling branch can learn semantic information which can provide extra cue for the study of single character. Ablation experiments show that these two branches have mutually reinforcing effects which can improve the total performance a lot.

\subsubsection{\textbf{Transcription Layer}}

In the transcription layer, we use CTC\cite{Graves2006Connectionist} as decoder to generate final output. Without predefined lexicon, CTC will simply output the character with the highest probability at each time stamp, concatenate them together and then remove the repeated ones and blanks. The transcription procedure can be denoted as:
\begin{equation}
\mathit{l}=\mathcal{B}({\arg\max}_\pi p(\mathbf{\pi}|\mathbf{y}))
\end{equation}
where 
$p(\mathbf{\pi}|\mathbf{y})=\prod\limits_{t=1}^Ty_{\pi_{t}}^{t}$, $y_{\pi_{t}}^{t}$ is the probability of outputting $\pi_{t}$ at time stamp $t$, $T$ is sequence length,  $\mathbf{y}=(y^{1}_{\pi_1},y^{2}_{\pi_2},...,y^{T}_{\pi_T})$ is the input sequence and $\pi = (\pi_1,\pi_2,...,\pi_T)$ is the output sequence. $\mathcal{B}$ is a mapping function which removes the repeated characters and blanks. $\mathit{l}$ is the final result. Due to the simple procedure, the computation cost of transcription layer is trivial.

\subsection{Loss Function}
%The entire network is supervised by two perspectives and loss comes from two branches. 
The objective function can be interpreted as follows:%which is a multi-task loss:
\begin{equation}\label{key}
\mathcal{L}=\mathcal{L}_{context}+\lambda\mathcal{L}_{char}
\end{equation}

$\mathcal{L}_{context}$ is context-level modelling loss and $\mathcal{L}_{char}$ is character-level modelling loss  from supervision enhancement branch.  Both of them are CTC loss.  
$\lambda$ is a hyperparameter which is used to trade off the weight of these two branches. $\lambda$ is set to $0.1$ in DSAN.

\section{EXPERIMENTS}
\subsection{Datasets}

\textbf{IIIT 5K-Words} (IIIT5K)\cite{Mishra2013Scene} contains 2000 word images in train set and 3000 word images in test set. 

\textbf{ICDAR 2013}\cite{Karatzas2013ICDAR} (IC13) test set contains 233 scene images. Following\cite{Cheng2017Focusing}, 1015 word images are cropped. %We discard images that has non-alphanumeric characters and the resulting word test dataset contains 1015 text images. 

\textbf{Street View Text}\cite{Babenko2012End} (SVT) 
contains 249 street view images, 647 word images are cropped.

\textbf{Synthetic Word Dataset}\cite{Jaderberg2014Synthetic} (Synth90k) contains 9 million synthetically generated word images. We used 800,000 images for model's pretraining.

\textbf{SynthText in the Wild}\cite{gupta2016synthetic} (SynthText) contains 800,000 synthetic images for scene text detection. We cropped 720,000 word images for model's pretraining.

\subsection{Implementation Details}
All the experiments are conducted on MXNet framework and the workstation has one GTX 1080Ti GPU.

%\textbf{Training}
We pretrain the model with 1.5 million synthetic word images, then finetune on each  benchmark. Each image is randomly rotated 0-3 degrees to the left and right respectively and then randomly zoomed out to 0.9-1.0 times of the original size. The image height is resized to 32 and the width is scaled according to the aspect ratio. Batch size is 32. The alphabet size is 37 with 26 English characters, 10 numbers and one blank. For BLSTM, 256 memory blocks are used. The learning rate is initialized as 0.1 and then divided by 10 after each epoch. Momentum optimizer is adopted. 

%\textbf{Inference}
%During inference, each input image is scaled as training time does. We use the transcription layer in context-level modelling branch to generate DSAN output.%We use multi-scale testing.

\subsection{Results}
Table \uppercase\expandafter{\romannumeral1} shows accuracy results of the proposed method on three benchmarks compared with some state-of-the-art methods. 
Baseline just use ResNet-34 based backbone and context-level branch without proposed text attention module and supervision enhancement branch. 
\begin{table}[htb]
	\vspace{-2pt} 
	\begin{center}
		\begin{tabular}{|c||c||c||c|}
			\hline 
			\diagbox{Method}{Dataset} &  IIIT5K & IC13 & SVT \\ 
			\hline 
			Photo-OCR \cite{Bissacco2013PhotoOCR} & \textendash & 87.6 & 78\\  
			CharNet \cite{Jaderberg2014Deep} & \textendash & 82.4 & 71.7\\
			CRNN \cite{Shi2017An} & 78.2 & 86.7 & 80.8\\ 
			RARE \cite{Shi2016Robust} & 81.9 & 88.6 & 81.9\\  
			STN-OCR \cite{Bartz2017STN} & 86 & 90.3 & 79.8 \\ 
			$R^{2}$AM \cite{Lee2016Recursive} & 78.4 & 90.0 & 80.7\\  
			Jaderberg et al. \cite{Jaderberg2014Reading} & \textendash & 90.8 & 80.7\\  
			Baidu IDL \cite{BaiduIDL} & \textendash & 89.9 &\textendash \\
			AON \cite{AON} & 87.0  & 91.5 & 82.8 \\
			\hline 
			baseline & 83.3 & 88.9 &75.7 \\
			%DSAN (simple) & 84.7 & 88.8 & 78.8 \\ 
			\textbf{DSAN} & \textbf{88.6} & \textbf{92.3} & \textbf{84.1\textbf} \\ 
			\hline 
		\end{tabular} 
	\end{center}
	\caption{\label{tab:1}Accuracy Results. All the results do not use any lexicon.  
	}
	\vspace{-12pt}
\end{table}

\vspace{-2pt}

In the entire framework, there are two transcription layers. For inference, we use transcription layer from context-level modelling branch as the output of DSAN. The last row in Table \uppercase\expandafter{\romannumeral1} is the results of DSAN. The accuracy results reach 88.6\%, 92.3\% and 84.1\% on IIIT5K, IC13 and SVT respectively which substantially outperform state-of-the-art methods. %which demonstrates the effectiveness of DSAN. 

\vspace{-2pt}

\subsection{Ablation Experiments}
In this section, we do two ablation experiments to exhibit the performance improvements brought by our key components: supervision enhancement branch and text attention module. 
\subsubsection{\textbf{The Effect of Supervision Enhancement Branch}}
In our method, $\lambda$ controls the weight of context-level modelling branch and supervision enhancement branch. Table \uppercase\expandafter{\romannumeral2} shows the effect of $\lambda$ on accuracies. In this part, we discard text attention module and just use plain ResNet-34 based backbone.

\begin{table}[h]
	\vspace{-3pt}
	\begin{center}
		\begin{tabular}{|c||c||c||c|}
			\hline 
			\diagbox{$\lambda$}{Dataset} &  IIIT5K & IC13 & SVT \\ 
			\hline 
			0 & 83.3 & 88.9 & 75.7\\ 
			\hline 
			0.05 & 87.5 & 91.2 & 81.9\\ 
			\hline 
			0.1 & \textbf{88.3} & \textbf{92.0} & \textbf{82.4}\\ 
			\hline 
			0.15 & 86.4 & 90.5 & 80.5\\ 
			\hline 
		\end{tabular} 
	\end{center}
	\caption{Accuracy results on three benchmarks with different $\lambda$.}
	\vspace{-16pt}
\end{table}

$\lambda=0$ means character classifiers are not working which is equivalent to no supervision enhancement branch. From the results, we can see that supervision enhancement branch can improve overall performance a lot and the whole network achieves relatively high performance with $\lambda=0.1$. Compared to $\lambda=0$, accuracy was improved 5\% on IIIT5K, 3.1\% on ICDAR2013 and 6.7\% on SVT which indicates the effectiveness of supervision enhancement branch. 

Fig.5 shows some real images that are identified wrong without supervision enhancement branch, but recognized totally correct with $\lambda=0.1$.

\begin{figure}[htbp]
	\vspace{-13pt}
	\centering
	\includegraphics[height=1.4in,width=3.55in]{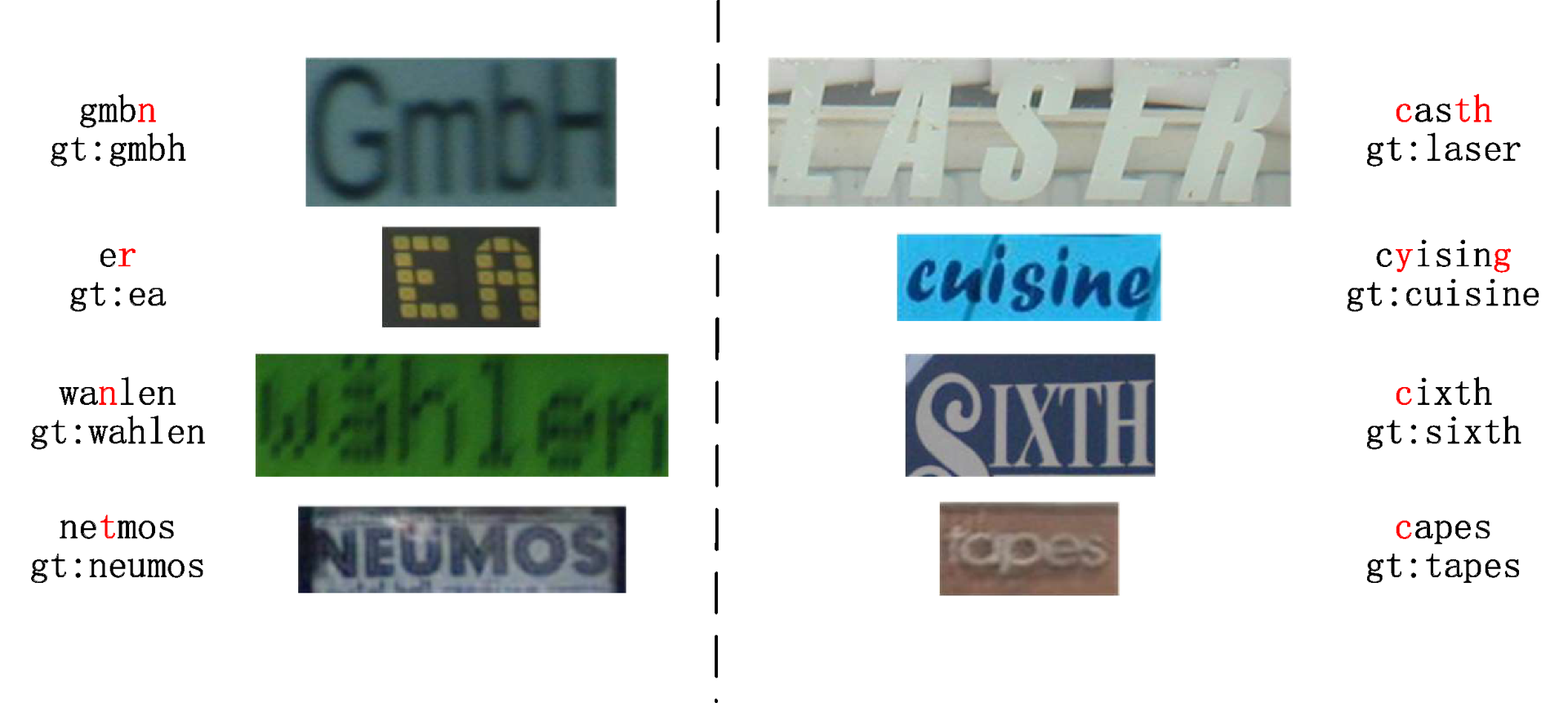}
	\vspace{-20pt}
	\caption{Red character represents incorrectly recognized character without supervision enhancement branch. gt stands for ground truth.}
	\vspace{-6pt}
\end{figure}

Texts in the left column do not have explicit semantic information which just are simple concatenation of characters while the right ones contain explicit contextual relationship between characters. It can be observed that the supervision enhancement branch not only enforces the recognition accuracy for text without explicit semantic information, but also further promotes the model's ability to capture contextual relationship.

\subsubsection{\textbf{The Effect of Text Attention Module}}
Table \uppercase\expandafter{\romannumeral3} shows the effect of text attention module. In this part, the weight of supervision enhancement branch $\lambda$ is set to 0.1.

\vspace{-3pt}

\begin{table}[h]
	\begin{center}
		\begin{tabular}{|c||c||c||c|}
			\hline 
			\diagbox{Method}{Dataset} &  IIIT5K & IC13 & SVT \\ 
			\hline 
			DSN & 88.3 & 92.0 & 82.4\\ 
			\hline 
			DSAN & \textbf{88.6} & \textbf{92.3}  & \textbf{84.1}\\ 
			\hline 
		\end{tabular}  
	\end{center}
	\caption{The effect of text attention module.}
	\vspace{-8pt}
\end{table}

\vspace{-5pt}

Compared to DSN (double supervised network without text attention module), accuracy was improved 0.3\% on IIIT5K, 0.3\% on ICDAR2013 and 1.7\% on SVT. Text attention module can improve the performance especially for SVT dataset which contains low-quality and challenging street view images. This is consistent with the intention of designing text attention module to suppress redundant features and clutters.
%------------------------------------------------------------------------- 
\section{CONCLUSIONS}

We present DSAN, an end-to-end trainable framework for scene text recognition which takes both explicit and inexplicit semantic information into consideration and incorporates one text attention module. It outperforms state-of-the-art methods on three benchmarks and can recognize text in arbitrary length without any predefined lexicon.

\section{Acknowledgment}
The research is supported by The National Key Research
and Development Program of China under grant
2017YFB1002401.

\vspace{-6pt}

\bibliographystyle{unsrt}
\bibliography{egbib}

\vspace{12pt}

\end{document}